# INDOOR GROUP ACTIVITY RECOGNITION USING MULTI-LAYERED HMM$_S$

## VINAYAK ELANGOVAN

Department of Information Sciences & Technology, Penn State Abington Abington, PA, 19001, USA

**Abstract** - Discovery and recognition of Group Activities (GA) based on imagery data processing have significant applications in persistent surveillance systems, which play an important role in some Internet services. The process is involved with analysis of sequential imagery data with spatiotemporal associations. Discretion of video imagery requires a proper inference system capable of discriminating and differentiating cohesive observations and interlinking them to known ontologies. We propose an Ontology based GAR with a proper inference model that is capable of identifying and classifying a sequence of events in group activities. A multi-layered Hidden Markov Model (HMM) is proposed to recognize different levels of abstract GA. The multi-layered HMM consists of N layers of HMMs where each layer comprises of M number of HMMs running in parallel. The number of layers depends on the order of information to be extracted. At each layer, by matching and correlating attributes of detected group events, the model attempts to associate sensory observations to known ontology perceptions. This paper demonstrates and compares performance of three different implementation of HMM, namely, concatenated N-HMM, cascaded C-HMM and hybrid H-HMM for building effective multi-layered HMM.

## I. INTRODUCTION

A wide variety of services offered by Internet of Things (IoT), for example smart homes, remote healthcare, transportation control, etc. rely on surveillance systems to process the collected data for making decisions and taking actions. Group Activity Recognition (GAR) is getting increasing attention as one of the most prominent challenge in Persistent Surveillance Systems (PSS). It requires the capability to recognize the actions of individuals and their connection to an activity being performed by a group. GAR requires the ability to process and recognize complex human interactions, and to identify the objective of such interactions. The main objective of PSS is to provide an early warning on potential threats while improving situational awareness in real-time. GAR is a challenging problem due to complexity involved in low-level processing, data alignment from different sensor sources, and fusion of soft and hard data, and interpretation of fragmented, but spatiotemporally correlated and associated information [1-3]. Human group activities may involve Human-Vehicle interactions (HVI), Human-Human interactions (HHI), Human-Object interactions (HOI), or multi-human-object/s interactions [2]. In this paper, HHI and HOI are considered in explaining GAR. HHI deals with identification of casual interactions between individuals in the group. Examples of HHI events are: greeting, shaking hands, waving hands, etc. HOI comprises type of interaction(s) that one may have with respect to certain object(s) used in a group activity. HOI represents the most challenging category for analysis since it is typically difficult to detect by the nature of object(s) involved. Examples of HOI events are: person carrying an object, person dropping an object, person placing an object, etc.

Reliability of HOI events detection through image processing becomes a difficult challenge due to objects occlusion and obstruction.

Matching behavioral pattern events of entity of interest with known ontologies helps in identification of Group Activities. For better understanding of this paper, the GA observations has been categorized into: "Action", "Activity", and "Group Activity". An Action refers to a simple operational pattern that a single or multiple entity may perform to change that state of the environment (e.g., a group walking, or person dropping an object etc.). The detected action of each entity is considered as an observation. An Activity refers to an array of associated and correlative actions that an entity may collectively perform to fulfill a specific task objective (e.g., a person placing an object on the floor). In general, this can be expressed as: Action$_1$ - Action$_2$ - Action$_3$ - …Action$_n$ $\rightarrow$ Activity$_{x1}$. For example, 'a person placing an object on the floor' activity is a collection of sequential actions such as: Person_carrying_object, Person_walking_with_object, Person_bending_down_with_object and Person_placing_object_on_floor $\rightarrow$ "Placing Object on floor" activity. A Group Activity refers to an array of spatiotemporally coupled activities that a group of entities may perform to achieve a specific common task objective (e.g., a team of individual cooperating for unloading a heavy object).

In the past, researchers have developed techniques and methodologies for better understanding of the information provided by imagery sensors. Researchers in [4] are able to classify group activities from video sequences by examining the actions of both the group as a whole as well as each individual member. They constructed a long short-term memory model using several layers of recurrent neural networks. Using this, they had success in identifying several group activities including walking, waiting, queuing, crossing, and talking. A research team in [5] was able to identify several human actions from video feed. They isolated specific actions using Motion History Image methods, and then were able to classify the actions being performed using Support Vector Machines binary classifiers. Using these methods, the researchers were able to successfully recognize actions such as walking, running, and waving. Researchers in [6] extracted silhouettes from videos and performed contour analysis to get outlines of the different poses of the test subjects while performing an action. They then clustered the different positions together to identify the sequence of actions that made up the activity. Finally, they used a Nearest Neighbor machine learning algorithm to classify different activities. They were able to recognize a variety of activities, including bending, jumping, walking, running and waving. However, they only used this method at a lateral angle to the subject, so it is unclear whether it would be as effective given video feed from a less optimal angle. Researchers in [7] were able to identify action sequences by focusing on identifying and extracting motion features from the person involved, any object they interacted with, and the background. Using the additional varied features, they were able to determine more accurately what action was taking place even when the actions of the person involved was less clear by itself. In [8] 3D Convolutional Neural Networks system was used to extract features from video input of human actions, and then the features are fed through a Recurrent Neural Network to classify a variety of human actions such as waving, clapping, running, and jogging. Researchers in [9] were able to classify small group activities by tracking the individuals in the group and detecting their behavior and interactions. They used a Gaussian Process Dynamical Model as the algorithm to learn the activity being performed by the group based on the features. They were able to classify activities like walking together, fighting, chasing, splitting, and following.

In [10] agglomerative and decisive clustering were used to identify groups and potentially suspicious group activities such as loitering, flanking, and aggression. Researchers in [11] were able to classify a variety of human actions by first creating a skeletal joint approximation to get a simpler tracking of the movement. Then, using Linear Discrete Analysis and vector quantization, they identified the general positions for specific actions. Finally, they trained the data using a Hidden Markov Model. They were able to accurately recognize several series of actions which included motions like walking, sitting, standing, carrying, and waving.

Much previous works has been devoted to develop algorithms for recognition of sequential actions performed by individuals or identical actions performed by groups like: group walking, group fighting, group gathering, group approaching, group chasing and etc. However, identifying GA involving complex sequence of actions have not been well addressed by previous researchers. By proper characterization of human interactions, appropriate semantic messages can be generated to describe the attributes of activities taking place with their spatiotemporal significance. In this paper, we present three competing HMM model for GA activity recognition. The remaining part of this paper is organized as such: group activity discovery and recognition framework, HMM modeling architectures, results analysis, and conclusion followed by acknowledgements and references.

## II. GROUP ACTIVITY DISCOVERY & RECOGNITION (GADR) FRAMEWORK

Group activity detection in PSS starts with the detection, identification, and tracking of targets via processing of data from multi-modality sensors (e.g., surveillance cameras, acoustic sensors or any other active monitoring sensors). Figure 1 presents an abstract representation of our Group Activity Monitoring System. The system contains six sequences stages. Stage-1 - is intended for target detection and refinement. This stage is typically involved with low-level image processing techniques that facilitate elimination of noises, background segmentation, image enhancement via appropriate digital filters, and extraction of feature vectors that collectively represent content-dependent attributes of Targets of Interests (TOI) with low order of representation dimensionality. Stage-2 - involves with spatiotemporal identification and tracking of TOI based on their trajectories motion analysis. Stage-3 – is intended for detection of HHI and HOI events and semantically annotating the events. In this stage characterization of detected HHI/HOI is performed and their associated chronological actions are registered as sequential observations. To detect and recognize a group activity, the actions/events performed by each individual in the group are detected and recognized. These individual actions are subject to spatiotemporal association and correlation, hypothesis matching and activity sequence modeling to infer a group activity. Stage-4a – is devoted for matching the annotated HHI/HOI events in sequence against the known ontologies for adaptive scene messages generation. Stage-4b – is intended for detecting presence of anomalous behaviors from the detected HHI/HOI activities through state transition modeling of the detected events. Stage-5 - performs generation of semantic messages describing the details of the group activity. In stage-5, per each recognized group activity, an appropriate semantic message is generated based on a modified TML (Transducer Markup Language) data structure format. The composition of TML data structure format can be found in our reference [2] and not discussed here. The final stage is intended for Visual Analytics (VA) of multi-modality sensors and meant for further analysis and inference of generated TML messages revealing comprehensive nature of different group activities [1,2].

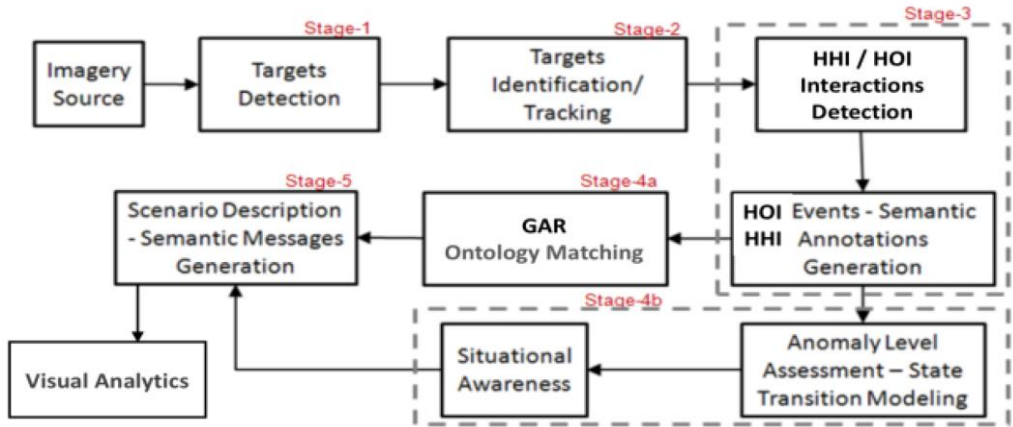

**Figure 1: Conceptual Overview of Group Activity Monitoring System**

## III. HMM MODELING ARCHITECTURES

HMM is a generative probabilistic model that can be trained to map observed sensory data onto latent hidden activity states. In the essence, the system being modeled is assumed to be markov process with unobserved hidden states. Mathematically an HMM can be expressed as: π, where A represents an (NxN) state transition matrix, B is an (NxM) observation probability matrix, and π is the initial state distribution. The activities (i.e., a sequence of recognized atomic actions) observed by the field sensors are coded and fed to an HMM for GA recognition. Training of HMMs are achieved primarily by constructing some ontology representing known group activities patterns. Through this process the HMM recognizes the likelihood of a test input sequence representing an activity, and maps that to a known trained ontology class with certain likelihood scale. Given the complexity of input sequences in terms of order and content, different implication may arise in interpreting the input sequence. To overcome this issue, we introduce three different competing architectures of Hidden Markov Model: Cascaded HMM, Concatenated HMM and Context-based HMM, and shown how to minimize such an uncertainty by combining the power of these three HMMs. The following sections describe the design concerns and the architectures for each model.

### 3.1 HMMs Activity Modeling Design Concerns

The HMM-based Group Activity Recognition system is developed for predicting the group activities from the traced evidential sequential observations. In principle, an HMM is a probabilistic model whose latest output depends only on the current state of the system. To assist GA-HMM to recognize a group activity it suffices to construct a sequence of observations implying that activity. We call such training sequences as "group activity ontology" or simply GA-ontology. Therefore, GA-ontology represents a list of unique observations where each observation implies a salient and unique group activity.

A typical group activity may short-lived or long-lived. Through visual image processing, each such activity may generate a limited or large number of observations. In order to improve predictability of GA-HMM to recognize the GA activities, there are many options available. One option is to input GA-HMM with a fix number of observations at a time. Another such option is feed the collective observations of a GA to GA-HMM as shown in Figure 2.

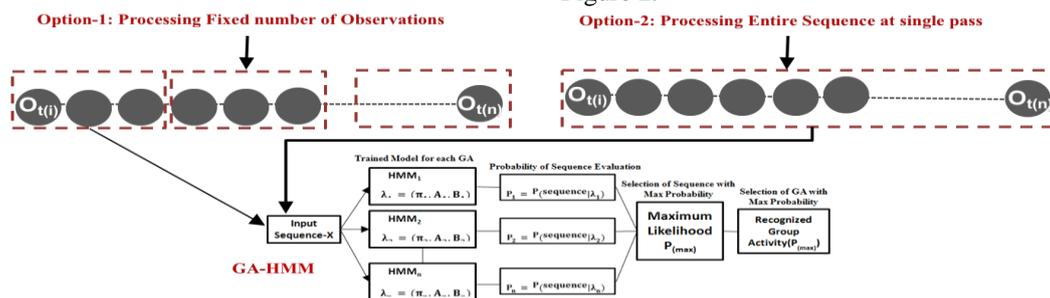

**Figure 2: Observation input sequence for GA-HMM**

For each evidential observation, it is assumed that there exists a state transition and an emission transition. For developing GA-HMM model, the following are the important parameters in developing HMM which affect the likelihood measure of recognized activity: number of states and observations, number of iterations or the convergence factor, transition and emission probabilities and initial probability measures. On the other hand, generating the input sequence and modeling the activity prediction using GA-HMM requires the consideration of different vital parameters. The following are the important parameters considered in generating the input observation sequence: 1. Spatiotemporal dependent events, 2. Number of events in a input sequence, 3. Association and correlation between events, 4. Uncertainty involved in events recognition and 5. Frequency of events/sequential bonded events GA Observations detected through image processing were labelled with associated Spatiotemporal and entity IDs. The developed GA-HMMs predicts the type of activity involved with emerging sequential observations. In this work, 90 salient observations were used for building the ontology for GA recognition including Human-Vehicle Interactions (HVI) [12]. An exemplary sample space of GA observations is shown below. In this list each observation is separated by a comma and refers to an action detected by the image processing technique in charging processing and annotating observed events. [Standing,Bending,Sitting,Laying_down,Walking,Running,Sneaking,Crawling,Waving_hands,Taken_from _cabinet,Person_Picked_Object,Person_Carrying_Object,Person_Dropped_object,Placed_Object_cabinet, Object_Left_Behind,Cabinet_open,………...,Shaking _hands,Group_Standing,Group_Walking,Group_Running,Group_Merging]

When more events in a sequence are fed to HMMs, the chance of neglecting an occurred activity is more since HMMs are efficient in predicting the current involved activity (state). The observation sequences are broken down to a desired window frame i.e. required number of events in a sequence, in order to avoid neglecting any activity being detected. The number of events used in constructing each HMM training ontologies is used as the ideal number of events in a sliding window frame of sequence. More than 100 ontology were developed for predicting 15 different kinds of GA. Table 1 shows a sample of developed ontology for Group Activity Recognition including HVI. As seen in the ontology Table 1, a maximum of three salient observations are used to construct each ontology pattern.

| Observation Sequence | Activity/Group Activity |
|---|---|
| *Person_walking_with_object, Person_bending_down_with_object and placing_object_on_floor* | Placing Object on floor |
| Object_picked-Object_Carrying- Object_placed_trunk | Loading-1 |
| *Object_carrying_object, Person_walking_with_object, and object_left_behind* | Object_Dropped |
| Object_taken_trunk-Object_Carrying-Walking | Unloading-1 |
| Entity-1_carrying_object-Entity-2_standing-Entity-2_carrying_object | Object_exchange-1 |
| Group_Merging-Group_United-Group_Shaking_hands | Social_interaction |

**Table-1: Activity/Group Activity Ontology**

As illustrated in Figure 1, a typical GA may be involved with multiple order of observations. The order by which these observations are inputted to HMMs matters.

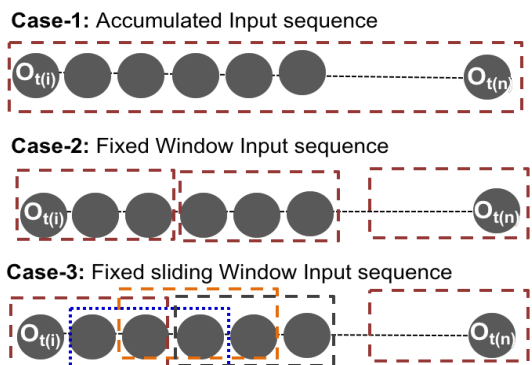

**Figure 3: Activity Input Observation Sequence**

In this study, we considered three different alternatives for grouping the observation input sequences. The three alternatives are shown in Figure 3.

If n is the number of events in a given observed sequence, w is the desired number of events in a input sequence to the HMMs, then the number of input sequence is:
Case-1: n÷w = 1 since n=w
Case-2: flooring(n÷w)
Case-3: (n-w+1)

In case-2, there may exist an instance where recent observations may be neglected. For example, if n is 11, w is 3, then the number of input sequence = 9. Two of the recent observations are neglected in input sequence generation. Also, some combinations of events in the sequence may not be in better correlation for predicting an activity.

Whereas, in case-3, since the sliding window is used to generate the sequences, all the events/observations are utilized in generating the input sequences. In both case-2 and case-3 more than one activity is predicted since the number of input sequences are greater than one, provided that () in case-2 and () in case-3. In our model, a Maximum Likelihood (ML) technique is used for the predicted activities in case-2 and case-3 for identifying the most relevant activity.

When generating the input sequence, it is essential to identify the association and correlation between events. For example in a sequence $(e_1-e_3-e_4)$ where $e_1$ and $e_3$ may not be associated or correlated with each other. There is a possibly in predicting an activity incorrectly. To perform the association and correlation between the events, equal weights are alloted to the ontology sequences. For example,consider an ontology $(e_1-e_3-e_5) \rightarrow$ Ontology-1, the associated weight for $(e_1 \rightarrow e_3)= 0.5$ and for $(e_3 \rightarrow e_5) = 0.5$. A cumulative weights between two events are calculated for all the ontology used in the activity prediction.

The accumulated weight of all correlated ordered pairs in an ontology = 1. Therefore the correlated weight of each pair is given by
In our work, the number of events in each ontology are equal. The cumulative correlated weights of each pair in all ontology are calulcated through the following algorithm.

```
Int Correlation_wt(int y)
  {
     If (y=1) return x
     Else return Correlation_wt(y-1) + ((1-
Correlation_wt(y-1))/2
  }
```
Let 'x' be the correlated weight of each pair and 'y' be the number of times a pair occurred in all the ontology. Correlation weights of each pair in the ontology is used in generating the input sequence. If the Correlation_wt is less than 'x', the combination of weight with the succeding event is calculated and the pair with the maximum weight is considered in the input sequence.

Devloped model accommodates to predict pertinent human activities by eliminating the events detected with less confidence. Removing trivial events and maintaining the distinct events in a input sequence also have significant impact in class (activity) prediction. Removing events of less importance i.e. trivial events helps in increasing the likelihood measures in appropriate HMM model selection. Spatiotemporal realtionship and frequency of events occurrence are also enforced in the activity detection. Certain operational activities can be recognized based on the frequency of occurrence of its sub activities. Detecting the frequency of occurrence with respect to time can modulate the severity of an activity or in another word, adjusts the level of Situtaion awareness (S_A) associated with a detected GA.

For eample if a person picks up an object, it may be recognized as if the person has removed an object and if a person drops an object in a space, it may be identified as if the person has placed an object. In our model, if event of 'Object Removed' and 'Object Placed' happens more frequent than the other, then the activity is identified as 'Loading' and 'Unloading' operations respectively, because of order frequency disparity between the two observated activties.

The above discussed design parameters are effectively considered in generating the input sequences for each experimental activity model. As mentioned earlier, three different model are used in activity prediction, namely, Cascaded model, Concatenated model and Context based model. The architecture of each model is discussed in the following sections.

### 3.2 Concatenated GA-HMM (N-HMM)
The main goal of the concatenated modeling is to build a generative model that estimates the most likely label at each input sample (i.e. window frame sequence). In the developed HMMs model, each hidden state is directly associated with a specific label i.e. group activity to be detected.

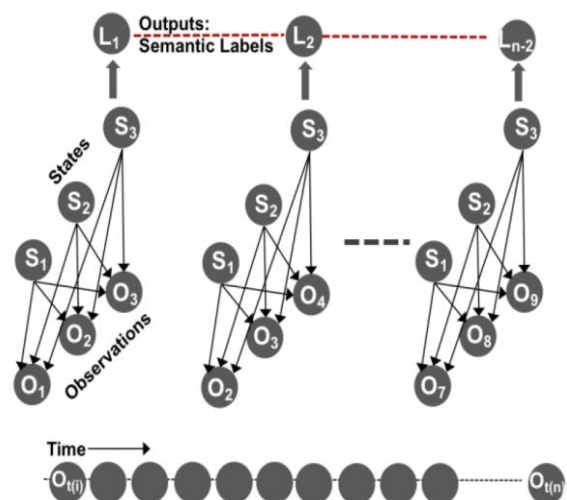

**Figure 4: Concatenated GA-HMM (N-HMM)**

For demonstration purpose, the number of events in an input window sequence is considered as three. The output labels recognized from each window sequence are fused together for further processing. The fusion strategy employed is discussed in the end of section-3. Figure 4 shows the model of Concatenated GA-HMM (N-HMM).

### 3.3 Cascaded GA-HMM (C-HMM)
In the cascaded GA-HMM (C-HMM), the inputs from the previous iteration of GA-HMM are fed forward to next iteration, namely, the output state from a preceding HMM is used an observation in input sequence for the next iteration. Figure 5 shows an example of a C-HMM where output from each iteration of GA-HMM is used as an observation input to next successive one.

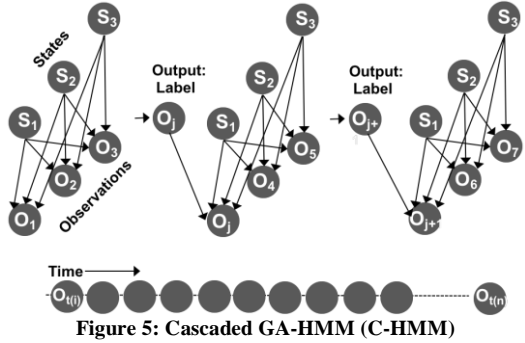

**Figure 5: Cascaded GA-HMM (C-HMM)**

The output from each GA-HMM is mapped with ontology for a possible match in an observation sequence. If the state is not considered in an input sequence, then model acts like N-HMM. Consider the following ontology:

Ontology-X1: Towards_cabinet - opens_cabinet - object_picked → Object_taken_cabinet
Ontology-X2: Object_taken_cabinet - Object_Carrying - Walking → Unloading

For example, consider the test sequence: [……Towards_cabinet-opens_cabinet-object_picked-Object_Carrying-Walking…].

In this given test sequence, HMMs uses Ontology-X1 and recognizes 'Object_taken_cabinet' activity and generated sequence is [……Object_taken_cabinet-Object_Carrying-Walking…]. The output label 'Object_taken_cabinet' is used as input in the next sequence, matched with Ontology-X2 for recognizing 'Unloading' activity.

### 3.4 Context-Based Hybrid GA-HMM (H-HMM)

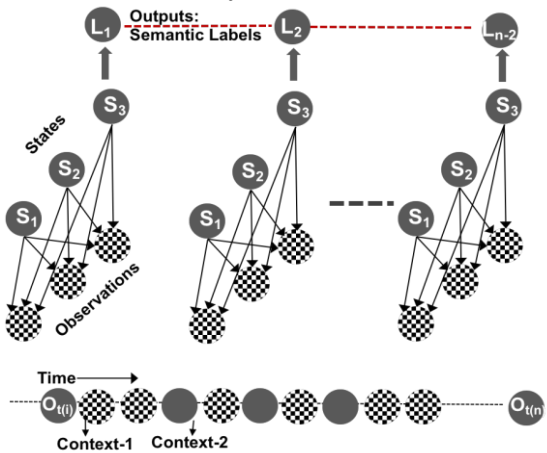

**Figure 6: Context based H-HMMs**

Though the identified observations are continuous i.e. sequential observations, the distributions of observations for generating the input sequences can be effectively modeled by a set of discrete classes independently for each context dimension. The test sequence is split into different input sequences based on the association among the events and a desired context as shown in Figure 6.

In a context of monitoring indoor activities (example: a classroom), events related to Human-Vehicle interactions are not relevant to predict GA. Therefore, depending on the context, an appropriate training ontology sequence would suffice to correct classify context-based group activities. Each model is selected based on the user requirements.

## IV. EXPERIMENTAL RESULTS

This section describes an experiment carried out for validation of group activity prediction in context of human-human and human-object interactions in a monitored environment. The following are the entities involved in the experiment: 3 humans, 1 midsized cardboard box and 1 small cardboard box. The tested "Group Exchanging Boxes" scenario shown in Figure 7 is as follows: Human-1 (H-1) and Human-2 (H-2) carry box-1 (O-1) walking together while Human-3 (H-3) is carrying box-2 (O-2) and waiting for H-1 and H-2; H-2 shakes hands with H-3; H-3 gives the O-2 to H-2; H-1 gives O-1 to H-3; H-1, H-2, and H-3 departs the scene. This scenario is modeled for GA recognition using the combination of the models described in previous section. For modeling this GA, the design criteria discussed in Section 3.1 are considered. A fixed window of input sequences as explained in section 3.1 is used to generate the input sequences for HMMs.

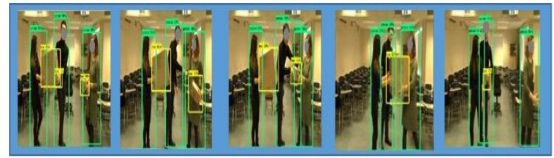

**Figure 7: "Group Exchanging Boxes" scenario**

A complete array of recorded observations from the scenario after spatiotemporally associating and correlating the events with their corresponding IDs are:

[H1_Walking,H1_O1_Object_carrying,H2_Walking, H2_O1_Object_carrying,H3_Standing,H3_O2_Object_carrying,Group_Merging,H1_Standing,H1_O1_Object_carrying,H2_Standing,H2_O1_Object_carrying, H3_Standing,H3_O2_Object_carrying,H2_H3_Hand shaking,H3_O2_Giving_H2,H2_O2_Receiving,H1_Standing,H1_O1_Object_carrying,H2_O2_carrying,H1_O1_Giving,H3_O1_Receiving,H2_Standing,H2_O2_Object_carrying,H3_O1_carrying,Group_Separating,H1_Walking,H2_Walking,H2_O2_Object_carrying,H3_Walking,H3_O1_Object_carrying,H1_Left,H2_Left,H3_left]

The observations are fed to three developed HMMs for predicting type of GA taken place in a classroom. In this case, actions performed by each entity are considered as the context. Table-2 shows the selected context and corresponding observation sequences.

| CONTEXT | OBSERVATION SEQUENCES |
|---------|----------------------|
| H1 actions. | H1_Walking, H1_O1_Object_carrying, Group_Merging, H1_Standing, H1_O1_Object_carrying, H1_Standing, H1_O1_Object_carrying, H1_O1_Giving, Group_Separating, H1_Walking, H1_Left |
| H2 actions | H2_Walking, H2_O1_Object_carrying, Group_Merging, H2_Standing, H2_O1_Object_carrying, H2_H3_Handshaking, H3_O2_Giving_H2, H2_O2_Receiving, H2_O2_carrying, H2_Standing, H2_O2_Object_carrying, Group_Separating, H2_Walking, H2_O2_Object_carrying, H2_Left |
| H3 actions | H3_Standing, H3_O2_Object_carrying, Group_Merging, H3_Standing, H3_O2_Object_carrying, H2_H3_Handshaking, H3_O2_Giving, H3_O1_Receiving, H3_O1_carrying, Group_Separating, H3_Walking, H3_O1_Object_carrying, H3_left |

**Table-2: Actions performed by each entity**

Each entity's observations are considered separately for recognizing their activity. For example, consider the actions performed by human H2 in Table-3.

| TASK PERFOMED | UPDATED OBSERVATION SEQUENCES |
|---------------|-------------------------------|
| H2 actions context Sequence for recognition (highlighted in bold) | H2_O1_Object_carrying, Group_Merging, H2_Standing, H2_O1_Object_carrying, **H2_H3_Handshaking, H3_O2_Giving_H2,** H2_O2_Receiving, H2_O2_carrying, H2_Standing, H2_O2_Object_carrying, Group_Separating, H2_Walking, H2_O2_Object_carrying, H2_Left |
| Result From: Cas-GA-HMM | H2_O1-Object_carrying, H2_O1_Object_carrying, H2_H3_Handshaking, H3_O2_Giving_H2, H2_O2_Receiving, H2_O2_carrying, H2_Standing, H2_O2_Object_carrying, Group_Separating, H2_Walking, H2_O2_Object_carrying, H2_Left |
| Sequence for recognition (highlighted in bold) | H2_O-1-Object_carrying, H2_O1_Object_carrying, **H2_H3_Handshaking, H3_O2_Giving_H2,** H2_O2_Receiving, H2_O2_carrying, H2_Standing, H2_O2_Object_carrying, Group_Separating, H2_Walking, H2_O2_Object_carrying, H2_Left |
| Result From: Con-GA- | H2_H3_Social_Interaction, H3_O2_Giving_H2, |
| HMM | H2_O2_Receiving, H2_O2_carrying, H2_Standing, H2_O2_Object_carrying, Group_Separating, H2_Walking, H2_O2_Object_carrying, H2_Left |
| Sequence for recognition (highlighted in bold) | **H2_H3_Social_Interaction, H3_O2_Giving_H2,** H2_O2_Receiving, H2_O2_carrying, H2_Standing, H2_O2_Object_carrying, Group_Separating, H2_Walking, H2_O2_Object_carrying, H2_Left |
| Results From: Cas-GA-HMM | H2_O2_carrying, H2_Standing, H2_O2_Object_carrying, Group_Separating, H2_Walking, H2_O2_Object_carrying, H2_Left |

**Table-3: Activity recognition for actions performed by Human H3**

This process is continued for all H2 observations. Methodology followed for recognizing activity performed by H2 is repeated for other entities for recognizing the GA in 'Group Exchanging Boxes' scenario.

## V. CONCLUSIONS

In this paper, we have shown that GA-HMMs can be effectively employed as a proper inference working model capable of analyzing information pertaining to observations extracted from video imagery frame by frame. The discussed architectures can be efficiently used for incremental perception of the GA as the observations emerge overtime. The proposed method may enhance group activity recognition in persistent surveillance systems thus improving performance of some IoT services. The focus of GA detection can be expanded further based on the requirements of the application to enhance a better characterization of group activities.

## ACKNOWLEDGMENTS

I thank Dr. Amir Shirkhodaie from Tennessee State University for his valuable guidance for this research development.

★ ★ ★